\documentclass[letterpaper,11pt]{article}
\PassOptionsToPackage{numbers, compress}{natbib}
\usepackage[preprint]{neurips_2022}

\usepackage{amsmath,amssymb,amsfonts}
\usepackage{algorithmic}
\usepackage{graphicx}
\usepackage{textcomp}
\usepackage{xcolor}
\usepackage{csquotes}
\usepackage{gensymb}
\usepackage{makecell}
\usepackage{algorithm}
\usepackage{booktabs}
\usepackage[utf8]{inputenc} 
\usepackage[T1]{fontenc}    
\usepackage{hyperref}       
\usepackage{url}            
\usepackage{nicefrac}       
\usepackage{microtype}      

\newtheorem{example}{Example}
\newtheorem{theorem}{Theorem}
\newtheorem{definition}{Definition}
\newtheorem{proof}{Proof}

\title{A Definition and a Test for \\Human-Level Artificial Intelligence}

\author{
  Deokgun Park \quad Md Ashaduzzaman Rubel Mondol \quad Aishwarya Pothula \quad  Mazharul Islam \\
  Computer Science and Engineering\\
  University of Texas at Arlington\\
  Arlington, Texas USA \\
  \texttt{deokgun.park@uta.edu} \\ \texttt{\{mdashaduzzaman.mondol, aishwarya.pothula, sxi7321\}@mavs.uta.edu } \\
}

\begin{document}

\maketitle

\begin{abstract}

  Despite recent advances of AI research in many application-specific domains, we do not know how to build a human-level artificial intelligence (HLAI). We conjecture that learning from others' experience with the language is the essential characteristic that distinguishes human intelligence from the rest. Humans can update the action-value function with the verbal description as if they experience states, actions, and corresponding rewards sequences firsthand. In this paper, we present a classification of intelligence according to how individual agents learn and propose a definition and a test for HLAI. The main idea is that language acquisition without explicit rewards can be a sufficient test for HLAI.  

\end{abstract}

\noindent There have been many ups and downs in artificial intelligence (AI) research, and many people made great advances in diverse applications, such as speech recognition, image recognition, game playing, or self-driving cars.   
Despite this, the limitation of the current state of the art is most apparent in the context of robotics. 
When a layperson or popular culture imagine AI, it is frequently associated with a butler robot that can do many services a human butler could provide. 
The robot would converse with other humans and robots to do more tasks. 
If someone asks for a new dish, it might search the Internet for a recipe and learn to prepare it. 
AI is thought as a software part for such a robot. 
It might be convenient if there is a specific name for such aspect for AI research,  because the term AI has a broader meaning nowadays. 
Alternative terms such as true AI, strong AI, or artificial general intelligence (AGI)~\cite{goertzel2007artificial} are often used, but they are not clearly defined.

In this paper, we suggest naming a sub-field of AI research for something like a butler robot as human-level artificial intelligence (HLAI). 
We provide a formal definition and a test as a theoretical common ground for the HLAI research. 
Specifically, we try to answer following questions. 

\begin{itemize}
    \item What is the verifiable or measurable difference between human intelligence and other animals? 
    \item What does it mean to learn with the language? 
    \item How can we test whether an agent has the HLAI?
    \item How can we administer such a test practically to aid the model development?
\end{itemize}

Let us begin by explaining what distinguishes the human-level intelligence from the rest.

\section{Level of Intelligence}
It would be helpful for our discussion to clarify a few terms such as intelligence, instinct, learning, language, and human-level intelligence. This will explain  why we promote a new term, ~\textit{HLAI}, instead of well-established terms, such as AI or artificial general intelligence (AGI). These definitions draw from an examination of biological actors - an earthworm, a rabbit, a monkey, and a human baby --- to distinguish different levels of intelligence.

Let us examine the nature of ~\textit{intelligence}  with a concrete  question of whether an earthworm is intelligent.
The answer will depend on the definition of intelligence. 
Legg and Hutter proposed the following definition for intelligence after considering more than 70 definitions from psychology and computer science~\cite{legg2007universal,legg2007collection}:

\begin{displayquote}
\textit{Intelligence} measures an agent's ability to achieve goals in a wide range of environments. 
\end{displayquote}

This definition is universal in the sense that it can be applied to a diverse range of agents such as earthworms, rats, humans, and even computer systems. Maximizing gene spreading, or ~\textit{inclusive fitness}, is accepted as the ultimate goal for biological agents~\cite{dawkins2016selfish}. Earthworms have light receptors and vibration sensors. They move according to those sensors to avoid the sun or predators~\cite{darwin1892formation}. These behaviors increase their chance of survival and inclusive fitness~\cite{hamilton1964genetical}. Therefore, we can say that earthworms are intelligent. 
If we agree that an earthworm is intelligent, then we might ask again if it has a \textit{general intelligence}. Considering that it does feed, mate, and avoid predators in a diverse environment,  it does have general intelligence. However, we would not be so interested in replicating an earthworm-like intelligence. That is why we suggest using HLAI as a term for our community's goal instead of more established terms such as artificial general intelligence (AGI).


However, there are differences in intelligence between earthworms and more advanced agents such as rats and humans. \textit{Behavior policy} is a function that maps a sensory input with the appropriate action. 
The behavior policy of an earthworm is hard-coded and updated only by evolution. In other words, it  is \textit{instinct}~\cite{tinbergen1951study} that is innate and does not change with experience. For rats and humans, the behavior policy does change with experience which is \textit{learning}. In this paper, we propose three levels of intelligence based on how learning is achieved in agents.  Table 1 shows a summary of this idea. 

\paragraph{Level 1 Intelligence}
In this categorization, earthworms have Level 1 intelligence, where there is no learning occurring at the individual level. Their behavior policy have a hard-coded mapping from sensory input to the corresponding action that is instinct updated with evolution~\cite{tinbergen1951study}. 

\paragraph{Level 2 Intelligence}
The problem with Level 1 intelligence is that the adaptation with evolution is slow. For example, if there is an abrupt climate change due to the meteor crash, agents with Level 1 intelligence will have difficulty adapting to the new environment in a timely manner. Furthermore, the behavior policy is encoded in the genetic code. If a species want to adapt to various environments such as diverse climates, the behavior policy has to be encoded in the genetic code, which is costly. If an agent can update behavior policy during its lifetime by \textit{learning} new rules such as a new type of food or shelter, it would increase the inclusive fitness and reduce the amount of the genetic code for diverse environments. 

 Let's call \textit{experience} as a sequence of sensory inputs (states) and agent actions. A reward is a special case of sensory input given by the internal reward system conditioned by the state.  We call those agents with the capability for learning with experience as \textbf{Level 2 intelligence}. 

To enable learning at the individual level, at least two functional modules would be required in addition to the level 1 agents. 
The first is a memory to store newly developed rules. The second module is a reward system to judge the merit of the state. We stated that the goal of a biological agent is to spread genes. However, the correct assessment is not possible during the life time of an individual agent.  For example, an agent may lay eggs in a hostile environment that no descendant will survive. Still, the agent cannot know this because it would perish long before this happens. Therefore, an agent with level 2 intelligence requires a function to estimate whether the current stimulus or state is good or bad during an agent's life.  The reward system serves this purpose by providing a proxy for the value of the state for the inclusive fitness.  

We point out that the environment does not provide a reward. Instead, it is an agent that produces a reward signal, which is the agent's estimate of the value of the current state. A dollar bill can be rewarding for some cultures but might not generate any reward for a tribal human who has never seen any money before. As for another example, when we eat three burgers for lunch, the reward for the first and third burger will be different, even though it is the same object for the sake of the environment. 

\begin{figure*}[tb]
  \centering
  \includegraphics[width=0.9\textwidth]{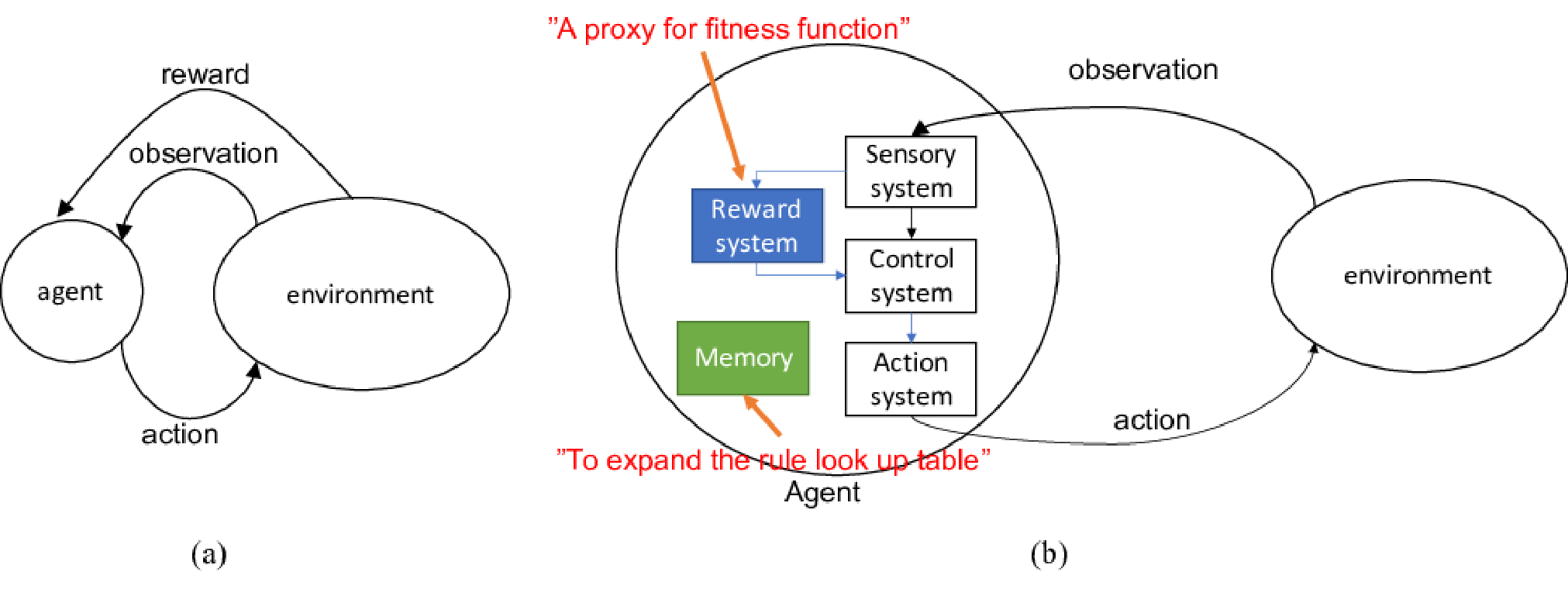}
  \caption{(a) The standard framework for reinforcement learning (b) The revised relationship of the agent and environment for level 2 intelligence. Environment provides an observation. Some observation is used for the reward system in the agent. The resulting reward signal and the sensory information is fed into the control system. New rules are added to the memory. }
  \label{fig:rl-framework}
\end{figure*}

However, this is different from the standard Markov Decision Process (MDP) framework for reinforcement learning, where a reward is determined from the environment. Legg and Hutter used a standard MDP framework for the formal definition of universal intelligence~\cite{legg2007universal}. However, they also commented   that a more accurate framework would consist of an agent, an environment, and a goal system inside the agent that interprets the state of the environment and rewards the agent appropriately.  

\paragraph {Level 3 Intelligence}
Contrary to our devotion to learning (machine, supervised, unsupervised, reinforcement, self-supervised learning, and so on), most behaviors of Level 2 intelligent agents are not based on learning but instincts.

\begin{example}
  Let us consider a rabbit that has never seen a wolf before.  If the
  rabbit tries to learn the appropriate behavior by randomly
  experimenting options when it does encounter a wolf, it is too late
  to update its behavior policy based on the outcome of random
  exploration.
  \label{example:rabbit_wolf}
\end{example}

 Instead, the rabbit should rely on the instinct which is the Level 1 intelligence.  Natural environments are too hostile to use learning as the primary method of building a behavior policy.   Therefore, the range of behavior policy  that Level 2 intelligence can learn with direct experience are limited.   
Level 3 intelligence overcomes this limitation by learning from others' experiences.

Bandura pioneered the  social learning theory  ~\cite{bandura1977social}, and learning through observation, thus called observation learning, is found on several species including non-human primates, invertebrates, birds, rats, and reptiles~\cite{ferrucci2019macaque}.  For example, if you give monkeys locked boxes that contain food, they will try to open them. When one monkey finds manipulation to unlock the box, other monkeys observe this pattern and imitate it to open their boxes.

\paragraph{Level 4 (Human-Level) Intelligence}  
The limitations of Level 3 which relies on the observation is also apparent. In the example ~\ref{example:rabbit_wolf}, the Level 2 rabbit relied on the direct experience. But for the Level 3 rabbit to learn the proper behavior, it has to observe its peer rabbit to be eaten by wolves which is also very rare event. Therefore, even Level 3 cannot learn a lot because they rely on the presence of the example case to be observed.

However, humans are the epitome of Level 3 intelligence and the only known species using language as a tool for social learning. 
The verbal and written language uses a sequence of abstract symbols to transfer knowledge, relieving the burdensome requirements of observational learning such as presence to demonstrations.  
Thus, we can think of human-level intelligence as \textbf{Level 3 intelligence with language}. 
In humans, a language is a tool for learning from others. 
Humans' technological achievements were possible because we can learn from others and contribute new knowledge. Isaac Newton said, ``If I have seen further, it is by standing on the shoulders of Giants.'' Language is an invention that enabled this. 
Verbal language enabled the knowledge transfer with the people at the same place and time. Later written language removed this barrier, and we don't have to be in the same place and time to learn from each other. 


 In the following sections, we will clarify the use of language for social learning because language has a various functions and forms.  



\begin{table}
\centering
\caption{Levels of intelligence}
\begin{tabular}{cl} 
\toprule
Level & \multicolumn{1}{c}{Features}                                                                                                                                                                           \\ 
\hline
1     & \begin{tabular}[c]{@{}l@{}}\begin{tabular}{@{\labelitemi\hspace{\dimexpr\labelsep+0.5\tabcolsep}}l}No individual learning\\Evolution-based refinement\\Ex) earthworms\end{tabular}\end{tabular}        \\ 
\hline
2     & \begin{tabular}[c]{@{}l@{}}\begin{tabular}{@{\labelitemi\hspace{\dimexpr\labelsep+0.5\tabcolsep}}l}Learning from direct experience\\Reward-based refinement\\Ex) rats, dogs\end{tabular}\end{tabular}  \\ 
\hline
3     & \begin{tabular}[c]{@{}l@{}}\begin{tabular}{@{\labelitemi\hspace{\dimexpr\labelsep+0.5\tabcolsep}}l}Learning from indirect experience\\Social, observation-based refinement\\Ex) primates, invertebrates, birds\end{tabular}\end{tabular}  \\
\hline
4 (Human-level) & \begin{tabular}[c]{@{}l@{}}\begin{tabular}{@{\labelitemi\hspace{\dimexpr\labelsep+0.5\tabcolsep}}l}Learning from symbolic experience\\Language-based refinement\\Ex) humans\end{tabular}\end{tabular}  \\
\bottomrule
\end{tabular}

\end{table}

\section{Clarifying Language Skill}
We need to clarify what we mean by learning with language. For example, dolphins are known to use a verbal signal to coordinate~\cite{janik2013communication}. Monkeys have been taught sign language~\cite{arbib2008primate}. Again, as we explained in previous sections, monkeys do learn by observation and imitations~\cite{ferrucci2019macaque}. But can we classify the language behavior of monkeys as human-level?   Similarly, there have been many previous works that demonstrated various aspects of language skills. Voice agents can understand the spoken language and can answer simple questions~\cite{kepuska2018next}. Agents have been trained to follow verbal commands to navigate~\cite{hermann2017grounded,chaplot2018gated,chen2019touchdown,das2018embodied,shridhar2020alfred}. GPT-3 by open AI can generate articles published as Op-Ed in the Guardians~\cite{brown2020language,gpt32020}.  
Some models can do multiple tasks in language as evaluated in the GLUE benchmark or DecaNLP~\cite{wang2018glue,mccann2018natural}.  Models exhibit superior performance than humans in all categories except for the Winograd Schema Challenge~\cite{levesque2012winograd}, where models perform slightly less than humans  ~\cite{raffel2020exploring}.
Do these models have human-level intelligence? 

Using language has many aspects. In this paper, we claim that learning from others' experience is the language's essential function that differentiates humans' language use with other animals'.  We will explain this with a simple example and then formalize it in the context of reinforcement learning. 

\begin{example}
Let's say that you have never tried Cola before. Now for the first time in your life, you see this dark, sparkling liquid that somehow looks dangerous. You have a few available actions, including drinking or running away. Randomly you might select to drink. It tastes good. It rewards you. It is not surprising that sugar water is frequently used as a reward for the primates in animal psychology experiments. 
\end{example}

Now your behavior policy for the same situation has changed such that you will choose to drink it more frequently next time you see cola. It is the change in the behavior policy induced by direct experience. This is how agents with Level 2 intelligence learn.

How an agent with Level 3 intelligence will learn?  Primates such as
gorillas and chimpanzees have Level 3 intelligence.  It means they can
learn from indirect experience.  They could learn by observing others
eat and the consequences.  Or a human teacher could point to a cola
glass and make an expression to make it attractive as a mother would
do to a baby.  In terms of the sequence of experience, they saw the
cola object and the response of other agents.

\begin{figure*}[tb]
  \centering
  \includegraphics[width=0.9\textwidth]{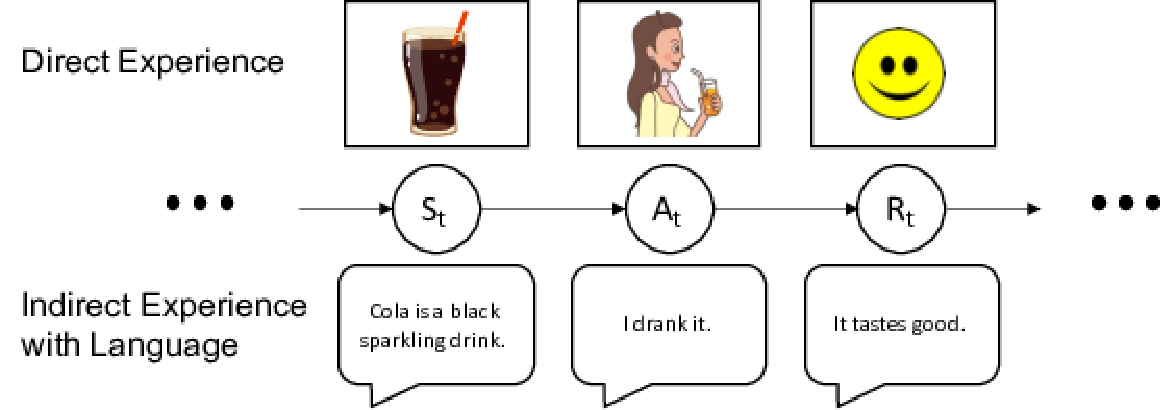}
  \caption{Learning with language means that the symbolic description brings the same changes to the model comparable to direct experiences.  }
  \label{fig:language}
\end{figure*}

Learning with language means that it should bring a similar change in your behavior policy when you hear someone say, ``Cola is a black, sparkling drink. I drank it, and it tasted good.'' Figure~\ref{fig:language} shows this with the notation in Markov decision process (MDP) (MDP)~\cite{sutton1998introduction}.   Humans use language for learning and this is what distinguishes a human-level intelligence from other animals with language.   
In this sense, we can define the human-level artificial intelligence (HLAI) as following;

\begin{definition}[\textbf{Human-level artificial intelligence (HLAI)}]
An agent has human-level artificial intelligence if  there exists  a sequence of symbols (a symbolic description)   for every feasible experience, such that the agent can update the behavior policy equally, whether it goes through the sequence of sensory inputs and  actions or it receives only the corresponding symbolic description.
\end{definition}

We can define more formally with Markov decision process (MDP).  Let $\mathcal S$ denote a set of all states, and $\mathcal A$ denote a set of all actions.  The stochastic behavior policy is given as $\pi(a|s) = p(a|s)$ where $a \in \mathcal A, s \in \mathcal S$. When the behavior policy, $\pi_{old}(a|s)$ is updated with a sequence of states and actions, $h$, we represent the updated policy as $\pi_{new} (a|s, h)$.  Given an original behavior policy, we can derive two policies $\pi(a|s, h_a)$ and $\pi(a|s,h_b)$ that are updated with two different experience $h_a$ and $h_b$.   We can measure the distance $Dist$ between two policies   using the expected KL divergence~\cite{schulman2015trust} w.r.t $s$.

\begin{equation}
  Dist (h_a, h_b) = \mathbb E_s [D_{KL}(\pi_a(a|s,h_a) || \pi_b(a|s,h_b))]
\end{equation}

Considering $s$ can be large, we might approximate the difference with the restricted set of states $s \in \mathcal S' \subseteq \mathcal S$, where we choose  $\mathcal S'$ to be relevant scenarios.  Let $\mathcal D$ represent the set of all sequences of states and actions that a biological agents can experience firsthand and $\mathcal T$ represent the set of all sequences of terms in language.

We might define a \textit{set of language} to aid the discussion as the following.

\begin{definition}[\textbf{A set of language}]
  A set of language is a set whose element is a tuple of an experience and a symbolic description, where the agent can update behavior policy equally either by going through the experience or by receiving the symbolic description.

  \begin{equation}
    \mathcal L = \left\{ (h_d, h_l) \in (\mathcal D, \mathcal T) |  Dist(h_l, h_d) \leq \delta \right\}
    \end{equation}
\end{definition}

In the previous example with Cola, the element for the language set
can be thought as the following.

\begin{itemize}
\item The direct experience is the sequence of the sensory stimulus.
\item The abstract symbol sequence is ``Cola is the black sparkling
  drink. It felt good when I drank it.''
\item The previous behavior policy is, given the black sparkling drink
  as the state, the agent might try evading or drinking it with equal
  probability.
\item The new behavior policy is that given the same state, the agent
  might try drinking it more likely.
\end{itemize}

Using a set of language, we can define HLAI as an agent with a language set, $\mathcal L_{human}$.

\begin{equation}
  ^\forall h_d \in \mathcal D,    ^\exists  h_l \in \mathcal T    \text{ s.t. } (h_d, h_l) \in \mathcal L_{human}
\end{equation}

It might lead to a philosophical debate whether a language set of human is indeed unbound. Authors claim that it is not bounded because it can be extended as needed.
In my definition,
human-level intelligence is defined with a language set for every
feasible experience which is infinite. But it does not mean that each
individual agent has to master a language set for infinite
experience. It is about the capability for handling open-ended
problems. For example, integer is infinite.  No human can see every
feasible integer in their lifetime. But when required, they can use
any of those integers even if they have never seen them before. As an
example with language, a typical English person will have only a few
words to describe shades of snow while an Eskimo might have more
words. But if a English person happens to spend 10 years with Eskmo
people, he might also acquire more language set for experience related
to snow.  Or how about a sentence ``He flew through the cheese
holes.''  Even though it is unlikely that someone has seen this sentence
before or experienced what the sentence describes, we have no
difficulty understanding the sentence or imagining some experience
that would justify the sentence.

Another example is how a fictional character, Scrooge in the novel,
\textit{A Christmas Carol}, might change the behavior policy with the
same verbal advice such as the virtue of the charity after experience.


However, one problem with implementing a test  according to this definition will be to make sure that there exists a symbolic description for every feasible experience.

\section{A Test for HLAI}

There are many tests for AI. However, a challenge is finding a sufficient but tractable one. There are many tests that are sufficient but intractable, including the Turing test, robot college student test, kitchen test, and AI preschool test~\cite{adams2012mapping}.   For example, the Turing test measures if an agent can imitate the human by communicating like one. 
The robot college student test asks an agent to register, take classes, and to get passing grades by doing assignments and exams.
Unfortunately, they are seldom conducted in the current research and when they are conducted, there is a controversy about the validity~\cite{shieber1994lessons}. 

There are a few limitations that make these tests impractical. First,  most tests assume that the agent has already acquired the language skill, but we do not know how to program an agent who can learn a language.   Second, they require human participants to administer the test. While it takes a few years for humans to be a master StarCraft II player, it took 200 years of gameplay for machines to masters~\cite{vinyals2019alphastar}.  Learning five years of human experience will take a lot of time for training with human intervention.  Therefore, using humans is cost-inhibitive and not scalable. Also, interactions with human participants are not reproducible for the validation.
Ideally, the test should require the minimum level of intelligence that can pass as human-level intelligence, and it should be cheap to run the test. 

At the other end of the spectrum, many tests for AI are tractable but not sufficient for  HLAI. While there are models with near-human or super-human level performance in Atari games~\cite{schrittwieser2020mastering}, Go~\cite{silver2017mastering}, Starcraft II ~\cite{vinyals2019grandmaster},  classifying objects from an image~\cite{he2016deep}, or multi-tasks in natural language understanding~\cite{he2020deberta}, none would claim that they achieved HLAI. They are effective in proposing a subset of necessary components or mechanisms for HLAI but are not built to study a sufficient set of those mechanisms. 

To find a Goldilocks middle ground between the sufficiency and tractability requirements,  we propose a new test for HLAI. If a human infant is raised in an environment such as a jungle where there are no human, he/she cannot acquire language. It is \textbf{environment-limited}. Also, if we have animal cubs and try to raise them like a human baby by teaching language, they cannot acquire language. It is \textbf{capability-limited}. Therefore, language acquisition is a function of an environment and a capability. 
Based on this argument, we propose the Language Acquisition Test for HLAI as the following; 

\begin{theorem}[\textbf{Language Acquisition Test  (LAT) for HLAI}]
Given a proper environment, if an agent with an empty set of language  can acquire a nonempty set of the language, the agent has the capability for HLAI. 
\end{theorem}

\begin{proof}
\textbf{(Proof by induction)} 
Suppose an agent can acquire a new element for the set of language that can bring the same change for a certain experience without relying on the existing set of language. In that case, the agent can keep adding elements to the set of language for a novel experience until it finds the symbolic description for any given experience.  
\end{proof}

\begin{example}
A baby will start learning a single word such as \textit{water} or \textit{mom}. When the baby hears these words, they bring similar effects such as seeing a cup of water or seeing mom. Even though this is a small start, the baby can continue to add the vocabulary to be fluent in the language. 
\end{example}

Compared to other tests, it has the small prerequisite. For example, the Turing test or robot college student test assumes that the agent has language skills, which is a challenging requirement for the current state of the art.

\subsection{Practical Administration of the LAT}
In the Language Acquisition Test, a proper environment means that there are other humans to teach language to the learning agent. 
A straightforward way to administer the test is by asking human participants to raise the physical robot agent like a human baby. Turing has suggested this approach~\cite{turing1950computing} and the Developmental Robotics community has actively pursed in many researches~\cite{lungarella2003developmental,asada2009cognitive,cangelosi2015developmental}. However, we already discussed the limitation of the human participants: the prohibitive cost and difficulty in reproducible research. 

It would be more useful if we could use a simulated environment~\cite{brockman2016openai}. There were previous works using simulated environments for the language acquisition, where agents get rewards by following verbal instructions in navigation~\cite{chen2019touchdown,savva2019habitat,chaplot2018gated,hermann2017grounded,shridhar2020alfred}  or give correct answers (question answering)\cite{das2018embodied}. 
 What is remarkable about these works is  the agent can understand the verbal instructions grounded to sensory input thus enabling compositionality of language. For example, let's say that an agent was trained to go to ~\textit{a small, red  box} and ~\textit{a  large, blue key} during the training phase.    As a result, during the test phase, the agent can successfully go to a ~\textit{a  small, blue box}, even though there was no such object during training.   However, previous environments have  following limitations for the test of the HLAI. 

 \begin{itemize}
    \item \textbf{Use of Rewards:} Using reward signals generated by environments will be sufficient for the implementation of Level 2 intelligence. However, for Level 3 intelligence, the reward is not given to the agent but is observed on other agents.  Similarly, for human-level intelligence,   the experiencing reward itself should be part of verbal description. In our previous cola example, there is a part related to the explicit reward that is ~\textit{it tasted good}. In the previous researches, they tend to use explicit reward to teach the concept of the ~\textit{black sparkling drink} by giving explicit reward when the agent point or navigate to the verbal description. ~\cite{chen2019touchdown,hermann2017grounded,chaplot2018gated,das2018embodied,shridhar2020alfred}. This approach cannot be applied in this case because we need a separate reward mechanism for teaching object concept ~\textit{black sparkling drink}  and the associated reward ~\textit{it tasted good}.  
    \item \textbf{Grounded Language and Embodied Exploration: }  The language symbols need to bring changes in the policy. It means that the language symbols need to be grounded with sensory input and the actions in the embodied agents. Some environments that use only the text lack this grounding.~\cite{narasimhan2015language,cote2018textworld}. 
    \item \textbf{Shallow interaction with large number of items and vocabulary: } Previous Environments tend to pour large items and vocabulary into the training. However, as Smith and Slone pointed out, human infants begin to learn a lot about a few things~\cite{smith2017developmental}. We need to build upon basic concepts before we can learn advanced concepts.  
\end{itemize}

Therefore, we claim that we need a new simulated environment for the test of HLAI to overcome these limitations.


\subsection{An Environment for Language Acquisition }


\begin{figure*}[tb]
  \centering
  \includegraphics[width=0.9\textwidth]{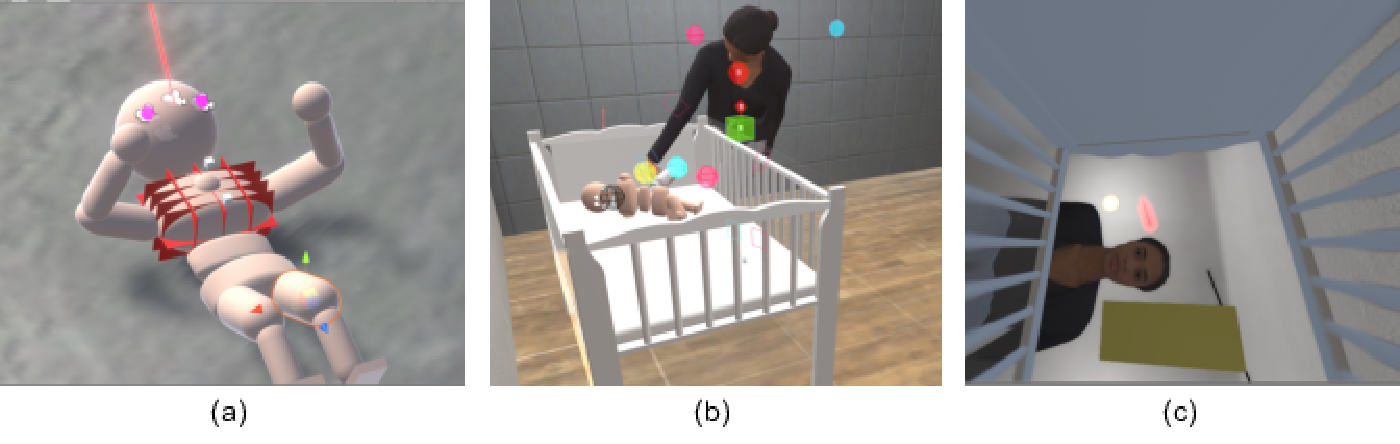}
  \caption{Screenshot of the SEDRo environment. (a) shows the learning agent which has the physical dimension of the one-year-old human baby. The orange line between eyes represents the eye gaze direction. The grid in the torso shows the area for the distributed touch sensors in the skin. (b) shows a caregiving agent feeds milk to the learning agent. (c) shows the visual input to the agent.   }
  \label{fig:sedro}
\end{figure*}


We have been working on Simulated Environment for Developmental Robotics (SEDRo) for the practical test of HLAI~\cite{pothula2020sedro}.  SEDRo provides diverse experiences similar to that of human infants from the stage of a fetus to 12 months of age.  
In SEDRo, there is a caregiver character (mother), interactive objects in the home-like environment (e.g., toys, cribs, and walls), and the learning agent (baby). 
The agent will interact with the simulated environment by controlling its body muscles and receiving  the sensor signals according to a physics engine. 
The caregiver character is a virtual agent. It is manually programmed by researchers using a behavior tree that is commonly used in video games to make a game character behave like a human in a limited way. 
Interaction between the agent and the caregiver allows cognitive bootstrapping and social-learning, while interactions between the agent and the surrounding objects are increased gradually as the agent enters more developed stages. 
The caregiver character teaches language by simulating conversation patterns of mothers. SEDRo also simulates developmental psychology experiments to evaluate the progress of intellectual development of non-verbal agents in multiple domains such as vision, motor, and social.  
The verbal speech is approximated by the sparse binary representations (SBR). Speech is encoded to a 512-dimensional vector, where about 10 of them are randomly selected for each alphabet. At each timestep, the corresponding speech signal is represented as the sequence of the vectors. 

SEDRo has the following novel features compared to previous works.
  
\begin{itemize}
    \item \textbf{Open-ended tasks without extrinsic reward}  In SEDRo, there is no fixed goal for the agent, and the environment does not provide any reward. Rather than relying on the environment for the rewards, the responsibility of generating rewards belong to the agent itself. In other words, AI researchers have to manually program a reward system to generate reward based on the current state. As an example, if an agent gets a food, the sensory input from stomach will change and  the reward system in the agent will generate a corresponding reward.  

    \item \textbf{Human-like experience with social interaction} Some studies use environments without explicit rewards, and the agents learn with curiosity, or intrinsic reward~\cite{singh2005intrinsically,bellemare2016countBasedExplorationIntrinsicMotivation}.  However, those environments were arbitrary and non-human such as robot arm manipulation tasks or simple games. While such simple environments are effective in unveiling the subset of necessary mechanisms, it is difficult to answer what is a sufficient set.  In SEDRo, we provide a human infant-like experience, because human infants are the only known example of agents capable of developing human-level intelligence. However,  we cannot replicate every aspect of human infants' experience, nor will we try to. There is a subset of experience that is critical for HLAI. Therefore, identifying  these essential experiences  and finding ways to replicate them in the simulation are two fundamental research questions.  Another benefit of a human-like environment is that we can use the experiments from developmental psychology to evaluate the development progress of non-verbal agents.

    \item \textbf{Longitudinal development} SEDRo unfolds agent capabilities according to a curriculum similar to human babies’ development. Many studies suggest that humans or agent models learn faster with constrained capabilities ~\cite{marklee2014longitudinalStudy, keil1981constraintCognitiveDev}. For example, in the first three months, babies are  very near-sighted and do not have any mobility. This makes many visual signals stationary, and the agent can focus on low-level visual skills with eyes. At later stages, when sight and mobility increase, babies can learn advanced skills built-upon lower level skills.  
\end{itemize}

The final benchmark whether the agent has acquired the language will follow the  protocol resembling the previous cola story. 
We give  verbal messages like \textit{``The red ball is delicious(good)''} or \textit{``The blue pyramid is hot (dangerous)''} and check if the behaviour policy toward \textit{the red ball} or \textit{the blue pyramid} has changed accordingly.  

\section{Discussion}

We proposed the definition and the test of HLAI. In this section, we discuss the implication of these on the current research on AI.   And we discuss the limitation of our approach and alternative options.

\subsection{Agent vs Behavior}

The levels of intelligence are to provide a novel insight on the research for artificial intelligence and not to provide new taxonomy for classification of biological agents. There are two limitations to apply this classification for the biological agents. First, we do not have a complete knowledge about intelligence of other animals. It is possible that later we might discover that  earthworms do learn new skills or other animals such as dolphins have more sophisticated use of language. In this case, we should adjust which animals belong which level of intelligence. A more fundamental second limitation is that biological species evolved for long times, and  boundaries tend to be blurry. 
For example, we might discern mammals from non-mammals with features such as laying eggs or not. But there is a platypus which is a borderline between mammals and reptiles~\cite{warren2008genome}. 
Similarly, there can be a gray area between what constitutes as the social learning with language.

Furthermore, the level of intelligence is better to classify behaviors rather than the biological agents. Higher level intelligence agents rely on the skills from the lower level intelligence.    For an  example, when a baby cries when hungry or shows stepping reflex, these behaviors are Level 1 intelligence. When they learn to avoid things after they experience pain, it is Level 2 behavior. Finally, when they observe and imitate the caregivers behavior with mobile phones, these behaviors are Level 3 in nature.

\subsection{Language border}

We claimed that humans are the only animals that are capable of learning with lanugage.  
Let us review  it with the hypothetical example of dolphins. 

\begin{example} 
  Let's say a dolphin says to other dolphin that ``There is a shark over the reef.'' Hearing this new information, the other dolphin might avoid the reef. 
\end{example}

In this case, it brought the change in the behavior but not in the behavior policy. In other words, dolphins have an innate behavior policy or instinct to avoid sharks. Hearing this information did not bring change in this policy. Let us differentiate a state information and an experience for our discussion. A state information refers the information about the state in MDP, while an experience refers to the sequence of states, actions, and rewards.  The message in this example was a state information because it was same as the other dolphin seeing the shark for itself. In other words, this verbal message is replacing the state in Figure ~\ref{fig:language}. In this case, we can say that  the verbal behavior of dolphins is not human-level considering language-based learning.  Greer et al. made a distinction between emission of a previously acquired repertoire and acquisition of a new repertoire~\cite{douglas2006observational}.  As a counter example, we might imagine dolphins doing the following conversation.  

\begin{example}

``I saw a fish with a shining string. I ate it. And there was painful experience.'' 
\end{example}

Hearing this message, if other dolphins avoid fishing bait, we can say these verbal behaviors are human-level intelligent behaviors. In this message, there are sequence of states, actions, and rewards and the behavior policy is updated with language. 

Again, we do not have the complete understanding of the language skills of dolphins and these examples are contrived. 
But the main purpose is to show the difference between the communication aspect and the learning aspect of the language. Communication is the sharing of a state information. Learning is when the behavior policy is updated with the verbal messages.
Probably, the language skills of advanced intelligent species such as dolphins and primates lie in the spectrum between two examples.

As long as I know, primates cannot learn with
purely abstract symbols.  That is the main point of the definition of
human-level intelligence.  Having said that it would not be surprising
if there is a case report that such learning is indeed possible in
primates probably with some simplification or blurry definition with
what an abstract symbol is.  Abstractness in language  means that the association of signified and signifier
are arbitrary~\cite{de1966course}.  However, there is a continuous spectrum in the
abstract symbol from explicit pictures to simplified iconographic
symbols to more abstract representation.  Some writing systems such as
Chinese characters still have some correspondence between symbol and
meaning.
Also, there is a continuous spectrum in intelligence, too.  After all,
primates like gorillas and chimpanzees are most similar to humans in
terms of intelligence. The difference between primates and humans will
be a matter of capacity than structure. For example, in terms of
computer architecture, personal computers running MS-DOS in the 1980s
and computers nowadays are very similar. It is just a matter of
capacities such as the size of RAM or the clock speed of CPUs. I
suppose the difference between primates and humans are subtle things
such as slightly more sophisticated control of the larynx or the
increased capacity of temporal sequence processing such as an
elongated hippocampal loop. Therefore, there must be an intersection
point between decreasing abstractness of symbols and increasing
intelligence of agents.


\subsection{Comparison with the Strong Story Hypothesis}
Patrick Henry Winston insisted that learning with language is the
essence of human-level intelligence. He posited 
\textit{the Strong Story Hypothesis}~\cite{winston2011strong}.  
  
\begin{theorem} [The Strong Story Hypothesis]
  The mechanisms that enable humans to tell, understand, and recombine
  stories separate human intelligence from that of other primates.
\end{theorem}

  \begin{example}
  As a friend helped me install a table saw, he said, ``You should
  never wear gloves when you use this saw.'' At first, I was
  mystified, then it occured to me that a glove could get caught in
  the blade. No further explanation was needed because I could imagine
  what would follow.
\end{example}

This is an example where a symbolic sequence has an effect similar  with a direct experience for the update of behavior policy.  Our contribution compared to his hypothesis is to formalize mechanically what does it
mean that human understand language.  In previous works, such understanding is formalized as text summarization, question and answering that can be ambiguous.  However, we claim that the essence of language understanding lies in
updating behavior policy.  The benefit of our definition is that it can be mathematically calculated using Markov decision process notation.  

\subsection{AGI or HLAI}

The history of AI is long, and the term AI is used in a broad sense.  While AI includes HLAI, it also includes active research area of application-specific AI or machine learning. Interestingly, when the general public thinks AI, they tend to think HLAI, while most academic research is on application-specific AI. Strong or True AI has been used to distinguish the two, but the definition is not clear. Artificial general intelligence (AGI) is also used in a similar context. AGI emphasizes that the agent should be able to do many things as humans do. However, doing many things in a diverse context does not necessarily mean that agents can do what humans do. As a counter-example, a rat can jump around, gather food, mate, and raise a newborn. A virtual rodent by Merel et al. can do multiple tasks depending on the context~\cite{merel2019deep}. We might say that this virtual rodent achieved AGI in the simulated environment, but this is not what AGI research targets.
As another example, humans can learn new alphabet from foreign language, but cannot learn to read QR code.  This shows that humans have also limited general intelligence.  This shows that biological agents have different degress of general intelligence.  But measuring the generality is not clearly defined or computationally intractable for practical cases.  In this sense, HLAI might be a better concept for AI research. 

\subsection{Merging instinct and learned behaviors}
Instinct is an umbrella word for innate behavior policy, and there are different implementation mechanisms including reflex, emotion, and special-purpose structures. For example, raising the arms when tripping, sucking, crawling, and walking are examples of reflex. Reflex relies on dedicated neural circuits. It is useful when it is okay that the response is rigid or fixed  and the reaction duration is instantaneous.  However, when a rabbit hears a wolf cry, the reaction needs to be flexible depending on the context. The reaction state should be maintained over longer time span. Emotion using neurotransmitters or hormones is effective in those cases, because its effect is global, meaning various areas of brain can respond according to it. And it lasts a long while before it is inactivated. Finally, the hippocampus or basal ganglia are special-purpose structures that solve a particular problems such as memory consolidation or decision among conflicting behavior plans~\cite{brown2004laminar}.

Instinct is a shortcut that enables a reasonable behavior policy in the life time of individual biological agents. Given infinite time, an agent with  learning capability might learn all the things that an intelligent animals can do without the help of instinct. But in reality, we saw that most of behaviors are  based on instinct in the example of the rabbit and the wolf. Another way of emphasizing the role of instinct is that primates have a few more social instincts than dogs~\cite{cangelosi2015developmental, lee2020grow}, and humans have just a few more language instincts than primates~\cite{pinker2003language}. While the volume of neocortex among dogs, primates, and humans are different, they play more or less a same role in those agents. Therefore, we need to build an artificial instincts  to program a HLAI.

Therefore,  we should add non-homogeneous special-purpose modules to the cognitive architecture for an organic mix of innate and learned behaviors. Current SOTA tends to be more homogeneous in its structure, emphasizing learning only. Again, contrary to our devotion to various forms of learning, most behaviors are based on instincts. 
Important questions are ``What instincts enable social interaction, knowledge learning, and language acquisition?'', ``How do those instincts work?'' and ``How can we merge the instinctive behavior and the learned behaviors?'' 
However, not all instincts of humans need to be replicated. Of special interests are instincts that enable human level intelligence such as knowledge instinct~\cite{livio2017makes}, social instinct, or language instinct~\cite{pinker2003language}. 
Below are the instincts that we conjecture as essential for HLAI.  

\begin{itemize}
    \item \textbf{Social instinct:} Innate behaviors such as face recognition, eye contact, following eye gaze, and  attending to caregivers are essential for the social learning~\cite{cangelosi2015developmental, lee2020grow}.
    \item \textbf{Knowledge instinct:} Intrinsic motivation or curiosity plays an important role in the knowledge acquisition~\cite{oudeyer2007intrinsic, singh2005intrinsically, bellemare2016countBasedExplorationIntrinsicMotivation, haber2018learning}. The prediction errors in the prefrontal cortex will generate a reward in the reward system. 
    \item \textbf{Decision system:} Artificial amygdala will determine the mode of the brain operation among 1) fight or flight, 2) busy without conflicts (Type I), 3) focus (Type II), 4) boring. At the boring state, the knowledge instinct is activated.  Additionally, the ~\textit{artificial basal ganglia} resolves conflict in multiple behavior options with the learning with reward~\cite{brown2004laminar}. As a concrete example, primates have a reflex that foveates to a moving object (pro-saccade). However, this reflex can be overridden by a training with rewards such that 1) participants maintain the gaze on the center fixation point even though there is a moving object (fixation task), 2) participants maintain the gaze until the fixation point disappears, and then foveate to the moving object (overlap task), or 3) participants maintain the gaze until the fixation point disappears, and there will be a fixed time interval between the disappearance of fixation point and onset of targets in a fixed location (gap task)~\cite{hikosaka1989functional}. Monkeys can be trained to foveate to moving target (pro saccade) if the fixation point is red, or to move eye in the opposite direction of the target (anti-saccade) if the fixation point is green, too~\cite{everling1999role}. 
    \item \textbf{Language instinct:} In addition to social instincts, language specific instincts are babbling, attention to voice-like signals and so on. 
\end{itemize}

\subsection{Limitations and Alternatives of the Test}
We proposed to use human-like experience to teach language. 
The main challenge is that it is difficult to program the caregiver character to enable diverse but reasonable interaction with the random behaviors of the learning agent. 
It is expected to teach a few first words if we are successful. 
Some alternatives include using a completely artificial environment that is not relevant to human experience but still requires skills in many domains. 
For example, emergent communication behaviors that can be thought of as language have been observed in the reinforcement learning environment with multiple agents~\cite{eccles2019biases,cao2018emergent,das2018tarmac,foerster2016learning}. 
While we might find the clues about the learning mechanism, it might be challenging to apply to the human robot interaction because language is a set of arbitrary symbols shared between members~\cite{kottur2017natural}.

Another possibility is to transform existing resources into a learning environment. 
Using Youtube videos to create a diverse experience can be an example. 
However, Smith and Slone pointed out that those approaches use shallow information about a lot of things, while human infants begin to learn a lot about a few things~\cite{smith2017developmental}. 
Also, visual information from the first years consists of an egocentric view, and the allocentric view emerges after 12 Months. 
Another aspect is that humans learn from social interaction. 
While infants can learn language from having a Chinese tutor in the meeting, but they cannot learn by seeing the recorded video of tutoring~\cite{kuhl2007speech}.  
Therefore, we assume that we need to acquire necessary skills before we can learn from those sources.

\section{Conclusion}
In this paper, we propose a definition of HLAI. This definition emphasizes that humans can learn from others' experiences using language. Based on this definition, we proposed a language acquisition test for HLAI. A version of this test can be approximated by the simulated environment, and we hope that other researchers can use it to facilitate the research on HLAI.




\bibliographystyle{plainnat}
\bibliography{references}

\end{document}